\documentclass[10pt,twocolumn,letterpaper]{article}

\usepackage[pagenumbers]{cvpr} 
\definecolor{LightBlue}{HTML}{3498DB}
\newcommand{\projecturl}[2]{\href{#1}{\textcolor{LightBlue}{\texttt{#2}}}}

\usepackage{amsmath,amssymb,amsthm}
\usepackage{booktabs}
\usepackage{multirow}
\usepackage{makecell}
\usepackage{pifont}
\usepackage{siunitx}
\usepackage{colortbl}
\usepackage{array}
\usepackage{xspace}
\usepackage{subcaption}
\usepackage{enumitem}

\newcommand{\method}{GS-IntSeg\xspace}
\newcommand{\cmark}{\textcolor{ForestGreen}{\ding{51}}}
\newcommand{\xmark}{\textcolor{Red}{\ding{55}}}
\newcommand{\best}[1]{\textbf{#1}}
\newcommand{\secondbest}[1]{\underline{#1}}

\DeclareMathOperator{\family}{fam}

\definecolor{tablehead}{HTML}{E7E9E8}
\definecolor{tablegroup}{HTML}{F3F4F3}
\definecolor{oursrow}{HTML}{EAF1EC}
\newcolumntype{Y}{>{\centering\arraybackslash}p{0.72cm}}
\setlist{topsep=2pt,itemsep=2pt,parsep=0pt}

\definecolor{cvprblue}{rgb}{0.21,0.49,0.74}
\usepackage[pagebackref,breaklinks,colorlinks,allcolors=cvprblue]{hyperref}

\def\paperID{13} 
\def\confName{3DV\xspace}
\def\confYear{2027\xspace}

\title{Do Gaussian Scenes Contain Enough Structure for Intrinsic Segmentation?}

\author{%
\begin{minipage}[t]{0.95\textwidth}\centering
Mohamed Rayan Barhdadi$^{1*}$\qquad Hasan Yazar$^{2*}$\\[2pt]
Erchin Serpedin$^{1}$\qquad Mehmet Tuncel$^{2}$\qquad Hasan Kurban$^{3}$\\[8pt]
{\normalsize $^1$Texas A\&M University\qquad $^2$Istanbul Technical University\qquad $^3$Hamad Bin Khalifa University}\\[8pt]
{\normalsize Project Page: \projecturl{https://kurbanintelligencelab.github.io/gs-intseg/}{kurbanintelligencelab.github.io/gs-intseg/}}
\end{minipage}
}

\begin{document}
\makeatletter
\let\gsintseg@maketitle\@maketitle
\def\@maketitle{%
    \gsintseg@maketitle
    \vspace*{-2em}
    \begin{center}
        \includegraphics[width=0.95\textwidth]{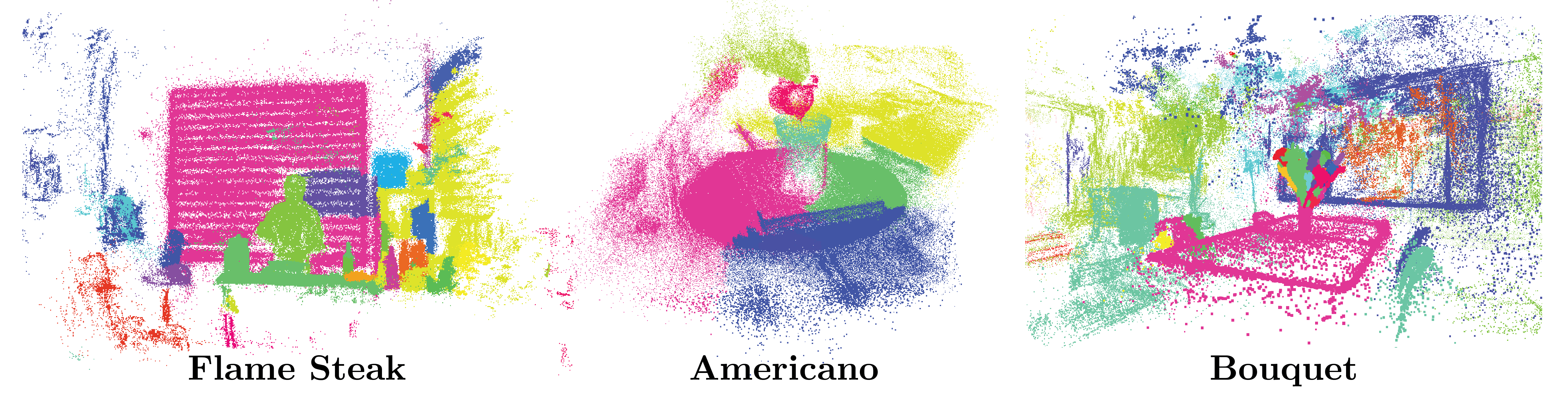}
        \captionof{figure}{\textbf{\method: intrinsic Gaussian
        segmentation without 2D masks.} The three-scene teaser shows
        flame\_steak from Neu3D on the left, americano from HyperNeRF in the
        center, and bouquet from LERF on the right. Region identities are recovered
        directly from the frozen Gaussian representation.}
        \label{fig:teaser}
    \end{center}
    \vspace*{0.5em}
}
\makeatother
\maketitle

\renewcommand{\thefootnote}{$*$}
\footnotetext{\raggedright Denotes equal contribution. Preprint. Correspondence to \texttt{rayan.barhdadi@tamu.edu}, \texttt{yazar22@itu.edu.tr}, \texttt{hkurban@hbku.edu.qa}.\par}
\renewcommand{\thefootnote}{\arabic{footnote}}
\setcounter{footnote}{0}
\begin{abstract}
Gaussian segmentation is usually posed as transferring object knowledge from 2D foundation models into a 3D representation. This leaves a  fundamental question unanswered: how much object structure is already encoded by a trained gaussian scene? We investigate this question with \method, an intrinsic, mask-free, and training-free method that constructs partitions using only: gaussian geometry, opacity, spherical-harmonic radiance, and deformation trajectories. On the dynamic Neu3D and
HyperNeRF datasets, \method obtains a mean of 0.677 mIoU across multi-view and monocular scenes without masks, external features, or segmentation training. This intrinsic formulation also enables \method to require approximately 2.5 minutes per HyperNeRF scene on a consumer RTX 5080 to construct its partitions, over \(10{\times}\) faster than SAM-based methods that require mask generation, feature rendering, and other stages.
These results suggest that gaussians alone can approach mask-supervised  performance in gaussian scenes segmentation. While a gap remains between intrinsic and foundation model based methods, to our knowledge, \method is the first mask-free approach to gaussian segmentation, motivating a re-evaluation of the assumption
that gaussian scene segmentation must be based on external masks and pointing toward an alternative faster, more generalizable segmentation approach. 
\end{abstract}
\section{Introduction}
\label{sec:intro}

Gaussian Splatting represents a scene with explicit anisotropic primitives
whose geometry, radiance, and identity persist through rendering
\cite{kerbl2023gs}. Dynamic extensions add deformation while preserving those
identities, producing a structured 4D representation
\cite{wu2024gs4d,yang2024deformable3dgs}. Object-level scene understanding
requires identifying the 3D components that constitute the reconstruction.
Current Gaussian segmentation methods instead obtain object information
primarily from external 2D models: SAM, CLIP, or another foundation model
produces a 2D signal that is
lifted, distilled, or tracked in the Gaussian scene
\cite{kirillov2023segany,ravi2024sam2,carion2025sam3,ye2024gg,cen2024saga,ji2024sa4d,li2026trase}.
This effective recipe leaves the representation itself largely untested and
makes segmentation performance, robustness, and consequently scene understanding and open-query
capability, heavily dependent on these external models, without even accounting
for the substantial computational cost involved in using them.

We ask whether Gaussian scenes contain enough intrinsic structure to support
segmentation before external semantics are introduced. Answering this
question separates information already present in the reconstruction from
information introduced by external supervision, clarifying what additional
semantic signals are actually necessary. We therefore treat segmentation
first as a question about the representation, not as another
feature-distillation problem.

A trained Gaussian scene provides several intrinsic grouping cues beyond
standard geometry-only point clouds. Covariance captures local surface
orientation and extent, opacity measures visible support, spherical harmonics
encode spatially varying radiance, and persistent Gaussian identity provides
motion trajectories in dynamic scenes. Together, these cues provide local
evidence for grouping Gaussians into coherent regions without semantic
descriptors or external segmentation masks.

\method converts these signals into a persistent partition bank over the
Gaussian scene. We first construct a sparse spatial graph over opaque
Gaussians. A material branch produces four fixed-scale partitions from DC-SH
appearance, covariance orientation, and Gaussian extent. In parallel, a
harmonic-flow branch constructs phase-sensitive and rotation-invariant
affinities from the full SH representation, combines them with geometric
continuation cues and reliable deformation common fate, and derives a nested
Infomap hierarchy. All partitions are computed directly from the frozen
Gaussian checkpoint and remain attached to canonical Gaussian identities.

For point-conditioned segmentation, a single positive point query selects one
candidate region from each frozen partition. We rank these candidates using
family-balanced agreement with a coverage penalty, then apply a
common-fate refinement in dynamic scenes to recover coherent moving extent
while controlling fragmentation. After the point query is fixed, the
evaluation mask is not used for partition selection, candidate ranking, or
refinement. The final prediction is therefore determined entirely by the
frozen Gaussian representation, the fixed partition bank, and the point query.

\begin{figure}[t]
\centering
\includegraphics[width=\columnwidth]{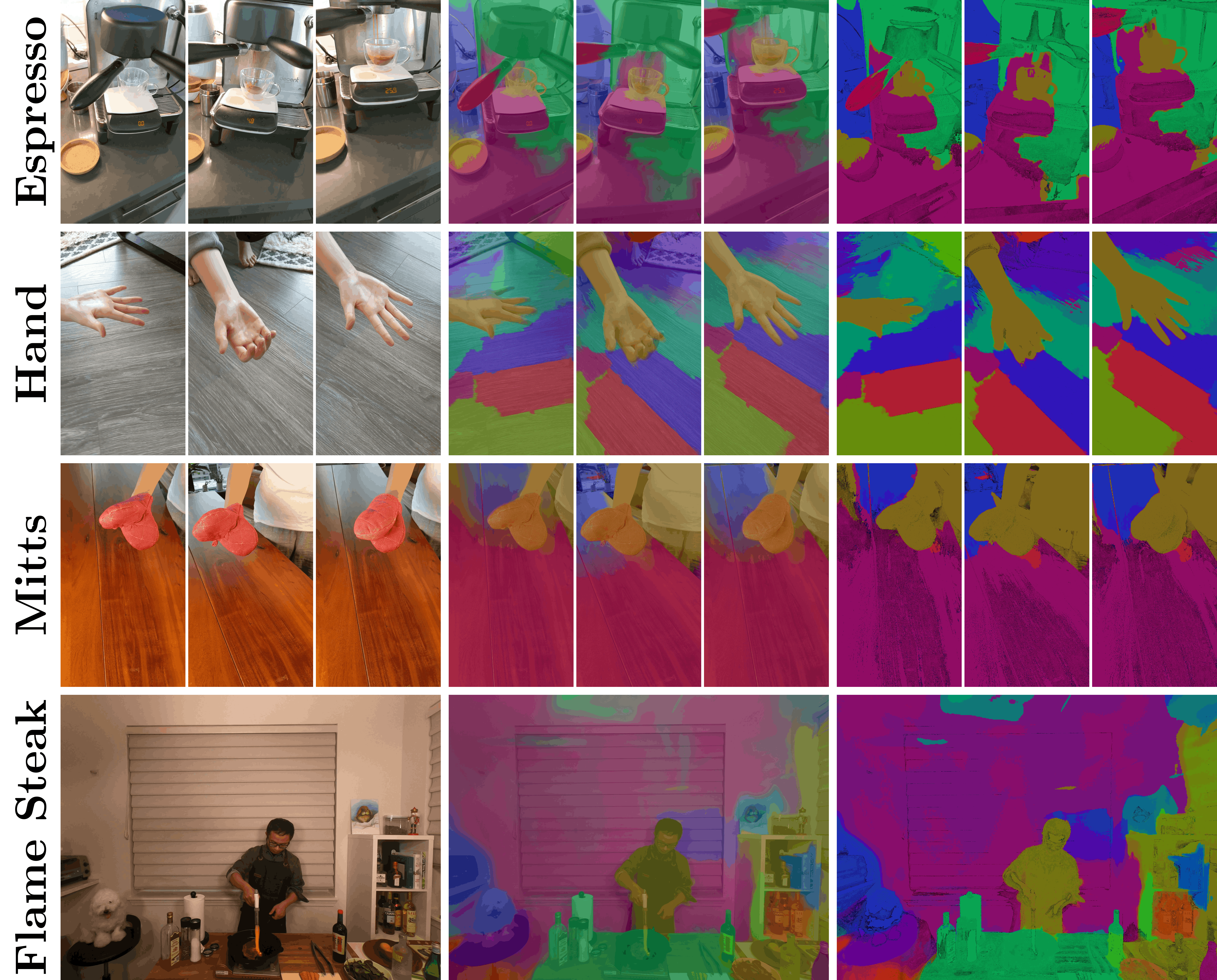}
\caption{\textbf{Qualitative dynamic segmentation.} For four dynamic scenes
(Espresso, Hand, Mitts from HyperNeRF, and Flame Steak from Neu3D), we show
three RGB frames (left), the corresponding persistent region overlays
(middle), and the categorical Infomap partition rendered without RGB
(right). Region identities remain attached to the same canonical Gaussians
across time, providing temporal consistency without per-frame mask
propagation. In Hand, the textureless floor is fragmented because the
intrinsic graph does not model a dedicated background class.}
\label{fig:dynamic-qualitative}
\end{figure}

Our contributions are:
\begin{itemize}
    \item an intrinsic Gaussian segmentation framework that uses only frozen
    reconstruction attributes, combining material partitions, a dual harmonic
    hierarchy, deformation-based common fate, and mask-independent
    point-conditioned consensus;
    \item a controlled evaluation on HyperNeRF and Neu3D in which ground truth
    is used only to place the reproducible point query and compute the final
    metric, reaching 0.899 mIoU on multi-view Neu3D and 0.566 mIoU on
    monocular HyperNeRF;
    \item a measured checkpoint-to-partition runtime of 150.3 seconds per
    HyperNeRF scene on a single consumer RTX 5080, \(15.0{\times}\) faster
    than TRASE's SAM mask-generation and feature-rendering stages while
    excluding TRASE's contrastive training.
\end{itemize}
\section{Related Work}
\label{sec:related}

\paragraph{Foundation-model Gaussian segmentation.}
Most Gaussian segmentation methods transfer object information from 2D
foundation models into the 3D representation. Gaussian Grouping
\cite{ye2024gg} and SAGA \cite{cen2024saga} use SAM-derived supervision to
learn object-aware Gaussian features, while Feature3DGS
\cite{zhou2024feature3dgs} and OpenGaussian \cite{wu2024opengaussian}
distill dense visual or language features into Gaussian fields. GaussianCut
\cite{jain2024gaussiancut} formulates interactive segmentation as graph cut,
and FlashSplat \cite{shen2024flashsplat} optimizes 2D-to-3D label transfer.
LUDVIG \cite{marrie2025ludvig} removes segmentation training but still
uplifts external DINOv2 features. In contrast, \method constructs its
partitions directly from frozen Gaussian geometry, radiance, opacity, and
deformation, without external masks or visual features.

\paragraph{Dynamic Gaussian segmentation.}
Dynamic methods extend 2D-supervised Gaussian segmentation across time.
SA4D \cite{ji2024sa4d} learns temporal identity features from tracked object
masks, while DGD \cite{bae2024dgd} distills DINOv2 or CLIP features into
dynamic Gaussians. TRASE \cite{li2026trase} removes explicit tracking but
learns a Gaussian feature field with SAM-mask supervision and contrastive
training before clustering. Split4D similarly couples dynamic reconstruction
with contrastive segmentation objectives for 4D scenes \cite{split4d2025}.
Language-grounded methods such as 4D LangSplat
\cite{li20254dlangsplat} and 4D Synchronized Fields
\cite{barhdadi2026_4dsync} similarly introduce external semantic features
into dynamic Gaussian representations. \method instead uses the frozen
reconstruction itself as the segmentation signal and requires neither mask
supervision, feature distillation, nor segmentation training.

\paragraph{Graph partitioning and intrinsic grouping.}
Graph-based segmentation groups nodes through local affinity, from normalized
cuts \cite{shi2000ncut} to modern community detection with Leiden
\cite{traag2019leiden} and random-walk compression with Infomap
\cite{rosvall2008maps}. Classical perceptual organization identifies
proximity, continuation, common fate, and boundedness as cues for object
grouping \cite{koffka1935gestalt,spelke1990principles,spelke2000core}.
\method instantiates these cues directly on Gaussian primitives: geometry and
covariance encode spatial continuation, spherical harmonics encode appearance,
and deformation trajectories encode common fate. Unlike prior Gaussian
segmentation methods, these affinities are computed entirely from the frozen
reconstruction.

\paragraph{Language-grounded Gaussian fields.}
Language-aware scene representations attach image-text features to radiance
fields or Gaussians to support open-vocabulary queries. LERF
\cite{kerr2023lerf}, LangSplat \cite{qin2024langsplat}, 4D LangSplat
\cite{li20254dlangsplat}, and LangSplatV2 \cite{li2025langsplatv2} embed
language-aligned visual features into the scene representation. These methods
target semantic querying and therefore rely on externally learned visual
representations. \method addresses the complementary question of how much
object structure can be recovered before such semantic features are added.
\section{Method}
\label{sec:method}

\subsection{Problem Formulation and Inputs}

Let a trained Gaussian scene contain primitives
\(
g_i=(\boldsymbol\mu_i,\mathbf\Sigma_i,\alpha_i,\mathbf H_i)
\),
where \(\boldsymbol\mu_i\in\mathbb R^3\) is the canonical center,
\(\mathbf\Sigma_i=\mathbf R_i\operatorname{diag}(\mathbf s_i^2)\mathbf R_i^\top\)
is the covariance with \(\mathbf s_i\in\mathbb R_+^3\),
\(\alpha_i\in[0,1]\) is opacity, and
\(\mathbf H_i=\{\mathbf h_{i\ell m}\}\),
\(\mathbf h_{i\ell m}\in\mathbb R^3\), are RGB spherical-harmonic (SH)
coefficients. Dynamic reconstructions additionally provide
\[
D_i(t)=(\Delta\boldsymbol\mu_i(t),\Delta\mathbf q_i(t),\ldots).
\]
Our goal is to assign persistent anonymous region IDs to the Gaussians and,
given a positive point query, return the corresponding object mask.

Partition construction uses only these checkpoint quantities. Camera parameters
are used only to render region IDs after partitioning. Input images, masks,
semantic labels, and external descriptors are not used to construct partitions.
Evaluation annotations are used only to place the deterministic point query and
compute metrics.

\begin{figure*}[t]
\centering
\includegraphics[width=0.9\textwidth]{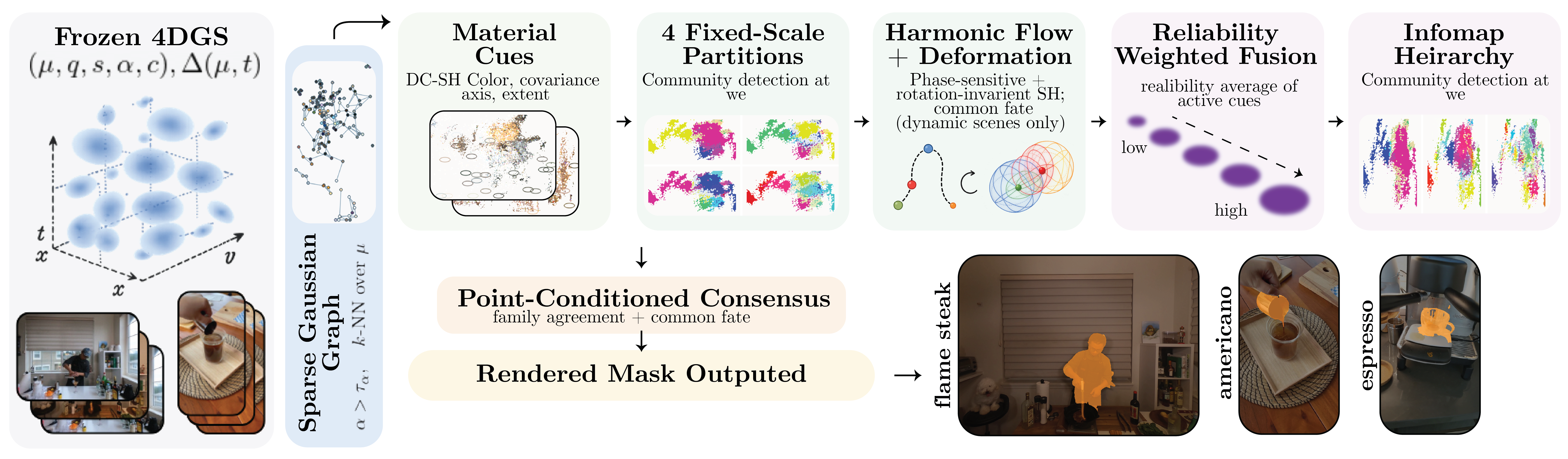}
\caption{\textbf{Pipeline overview.}
A frozen Gaussian scene supplies geometry, appearance, and, when available,
deformation cues to a sparse Gaussian graph. The material branch produces four
fixed-scale partitions, while harmonic and deformation affinities produce a
nested Infomap hierarchy. A positive point query selects a persistent region
through point-conditioned consensus and common-fate refinement. No 2D masks,
semantic features, or segmentation training are used to construct partitions.}
\label{fig:pipeline}
\end{figure*}

\subsection{Sparse Gaussian Topology}

We discard primitives with \(\alpha_i\leq0.05\) and connect each remaining
Gaussian to its \(k=20\) nearest canonical centers. Spatial proximity defines
candidate adjacency, while intrinsic affinities determine grouping.

The proposal bank contains two families: fixed-scale material partitions and a
nested hierarchy derived from higher-order SH and deformation cues.

\subsection{Material-Scale Gaussian Partitions}

For each spatial edge \((i,j)\), the material branch uses the DC-SH coefficient
\(\mathbf c_i=\mathbf h_{i00}\), principal covariance axis
\(\mathbf v_i^{\mathrm{axis}}\), and Gaussian extent \(\|\mathbf s_i\|_2\):
\begin{align}
A^{\mathrm{dc}}_{ij}
 &= \exp\!\left(
 -\frac{\|\mathbf c_i-\mathbf c_j\|_2^2}{2\sigma_c^2}
 \right),\\
A^{\mathrm{axis}}_{ij}
 &= \left|
 (\mathbf v_i^{\mathrm{axis}})^\top
 \mathbf v_j^{\mathrm{axis}}
 \right|,\\
A^{\mathrm{scale}}_{ij}
 &= \exp\!\left(
 -\frac{\log^2(\|\mathbf s_i\|_2/\|\mathbf s_j\|_2)}
 {2\sigma_s^2}
 \right),\\
w^{\mathrm{mat}}_{ij}
 &= \left(
 A^{\mathrm{dc}}_{ij}
 A^{\mathrm{axis}}_{ij}
 A^{\mathrm{scale}}_{ij}
 \right)^{1/3},
\label{eq:material-affinity}
\end{align}
with fixed \(\sigma_c=0.8\) and \(\sigma_s=1.0\).

We run weighted Leiden with the RBConfiguration partition at
\(\rho\in\{0.004,0.008,0.015,0.060\}\). Let
\(k_i=\sum_jw^{\mathrm{mat}}_{ij}\) and
\(M=\frac12\sum_i k_i\). At resolution \(\rho\),
\begin{equation}
Q_\rho(\pi)=
\sum_{ij}
\left(
w^{\mathrm{mat}}_{ij}
-\rho\frac{k_i k_j}{2M}
\right)
\mathbf 1[\pi_i=\pi_j].
\label{eq:rbconfiguration}
\end{equation}
All material parameters are fixed across scenes.

\subsection{Dual Intrinsic Harmonic Flow}
\label{sec:dihf}

The second family uses higher-order SH coefficients through complementary
phase-sensitive and rotation-invariant representations.

\paragraph{Phase-sensitive representation.}
For SH degree \(\ell\),
\begin{equation}
d^{\mathrm{phase}}_{ij,\ell}
=
\frac{1}{3(2\ell+1)}
\sum_{m=-\ell}^{\ell}
\|\mathbf h_{i\ell m}-\mathbf h_{j\ell m}\|_2^2.
\label{eq:phase-distance}
\end{equation}
Each band is normalized by its scene-median edge distance before averaging
across degrees.

\paragraph{Rotation-invariant representation.}
Stack degree-\(\ell\) coefficients as
\(\mathbf H_{i\ell}\in\mathbb R^{(2\ell+1)\times3}\) and define
\begin{equation}
\mathbf G_{i\ell}
=
\frac{\mathbf H_{i\ell}^{\top}\mathbf H_{i\ell}}
{2\ell+1}.
\label{eq:sh-gram}
\end{equation}
A common 3D rotation applies an orthogonal Wigner transform within a degree and
therefore leaves \(\mathbf G_{i\ell}\) invariant. For
\(c,d\in\{R,G,B\}\), \(c\neq d\),
\begin{align}
p_{i\ell c}
&=\sqrt{(\mathbf G_{i\ell})_{cc}+\epsilon}, \qquad
r_{i\ell cd}
=
\frac{(\mathbf G_{i\ell})_{cd}}
{\sqrt{(\mathbf G_{i\ell})_{cc}(\mathbf G_{i\ell})_{dd}+\epsilon}},\\
\mathbf z_{i\ell}
&=
[p_{i\ell R},p_{i\ell G},p_{i\ell B},
r_{i\ell RG},r_{i\ell RB},r_{i\ell GB}].
\label{eq:invariant-descriptor}
\end{align}

Let \(L\) denote the number of non-DC SH degrees. After robust
coordinatewise standardization, concatenate the degree descriptors into
\(\bar{\mathbf z}_i\in\mathbb R^{6L}\):
\begin{align}
d^{\mathrm{inv}}_{ij}
&=
\frac{\|\bar{\mathbf z}_i-\bar{\mathbf z}_j\|_2^2}{6L},\\
A^{\mathrm{inv}}_{ij}
&=
\exp\!\left(
-\frac{d^{\mathrm{inv}}_{ij}}
{2\max(
\operatorname{median}_{(p,q)\in E}d^{\mathrm{inv}}_{pq},
\epsilon)}
\right).
\label{eq:invariant-affinity}
\end{align}
Normalization and numerical implementation details are given in the
supplementary material.

\paragraph{Surface cues and reliability.}
Both harmonic branches combine their SH affinity with DC radiance, normalized
covariance shape, Gaussian support, local-frame SH appearance, and tangent-plane
continuation. For a \(d\)-dimensional descriptor \(\mathbf f_i\), let
\(\mathcal N(\mathbf f_i)\) denote robust scene-level coordinate normalization:
\begin{align}
d^f_{ij}
&=
\frac{\|\mathcal N(\mathbf f_i)-\mathcal N(\mathbf f_j)\|_2^2}{d},\\
A^f_{ij}
&=
\exp\!\left(
-\frac{d^f_{ij}}
{2\max(\operatorname{med}_E d^f,\epsilon)}
\right).
\label{eq:descriptor-rbf}
\end{align}

The DC cue uses \(\mathbf f_i^{\mathrm{dc}}=\mathbf c_i\). For covariance
shape, define
\[
\mathbf C_i=
\mathbf R_i
\operatorname{diag}\!\left(
\mathbf s_i^2/\|\mathbf s_i\|_2^2
\right)
\mathbf R_i^\top.
\]
Its descriptor contains the three diagonal entries and \(\sqrt2\) times its
three distinct off-diagonal entries of \(\mathbf C_i\).
Gaussian support and tangent continuation are
\begin{align}
d^{\mathrm{sup}}_{ij}
&=
\tfrac12\left(
\|\mathbf R_i^\top(\boldsymbol\mu_j-\boldsymbol\mu_i)/\mathbf s_i\|_2^2+
\|\mathbf R_j^\top(\boldsymbol\mu_j-\boldsymbol\mu_i)/\mathbf s_j\|_2^2
\right),\\
A^{\mathrm{sup}}_{ij}
&=
\exp\!\left(
-\frac{d^{\mathrm{sup}}_{ij}}
{2\max(\operatorname{med}_E d^{\mathrm{sup}},\epsilon)}
\right),\\
A^{\mathrm{tan}}_{ij}
&=
\left(
|\mathbf n_i^\top\mathbf n_j|
(1-|\mathbf n_i^\top\mathbf e_{ij}|)
(1-|\mathbf n_j^\top\mathbf e_{ij}|)
\right)^{1/3},
\label{eq:surface-cues}
\end{align}
where \(\mathbf n_i\) is the shortest covariance axis and
\(\mathbf e_{ij}\) is normalized center displacement. The local-frame cue
evaluates higher-order SH in the local surface frame and is converted to
\(A^{\mathrm{frame}}\) using Eq.~\eqref{eq:descriptor-rbf}.

For sorted scales \(s_{i1}\leq s_{i2}\leq s_{i3}\), define
\begin{equation}
\begin{aligned}
a_i^{\mathrm{ani}}
&=\frac{s_{i3}-s_{i1}}{s_{i3}},&
\nu_i
&=\frac{s_{i2}-s_{i1}}{s_{i2}},\\
o_{ij}
&=\sqrt{\alpha_i\alpha_j},&
h_i
&=\frac{e_i}{e_i+\operatorname{med}_{j:e_j>0}e_j},
\end{aligned}
\label{eq:node-reliability}
\end{equation}
where \(e_i\) is the RMS of the non-DC SH coefficients. With
\(\operatorname{OR}(x,y)=1-(1-x)(1-y)\) and
\(q_{ij}=o_{ij}\sqrt{h_i h_j}\),
\begin{align}
u^{\mathrm{sup}}_{ij}
&=1,&
u^{\mathrm{shape}}_{ij}
&=\operatorname{OR}(a_i^{\mathrm{ani}},a_j^{\mathrm{ani}}),\\
u^{\mathrm{tan}}_{ij}
&=\operatorname{OR}(\nu_i,\nu_j),&
u^{\mathrm{dc}}_{ij}
&=o_{ij},\\
u^{\mathrm{frame}}_{ij}
&=q_{ij}\operatorname{OR}(\nu_i,\nu_j),&
u^{\mathrm{harm}}_{ij}
&=q_{ij}.
\label{eq:cue-reliability}
\end{align}

For cue affinity \(A^r_{ij}\) and reliability \(u^r_{ij}\),
\begin{equation}
w_{ij}
=
\frac{\sum_r u^r_{ij}A^r_{ij}}
{\sum_r u^r_{ij}+\epsilon}.
\label{eq:reliability-fusion}
\end{equation}
No cue weights are learned or fitted per scene.

\paragraph{Deformation common fate.}
For dynamic scenes, we sample \(T=20\) times and center each translation
trajectory:
\begin{equation}
\widetilde{\boldsymbol\delta}_i(t)
=
\Delta\boldsymbol\mu_i(t)
-\frac{1}{T}\sum_{\tau}\Delta\boldsymbol\mu_i(\tau).
\end{equation}
Trajectory similarity is
\begin{equation}
A^{\mathrm{mot}}_{ij}
=
\max\!\left(
0,
\frac{
\sum_t
\widetilde{\boldsymbol\delta}_i(t)^\top
\widetilde{\boldsymbol\delta}_j(t)}
{\|\widetilde{\boldsymbol\delta}_i\|_F
 \|\widetilde{\boldsymbol\delta}_j\|_F+\epsilon}
\right).
\label{eq:motion-affinity}
\end{equation}

Let \(r_i\) denote the distance to the farthest retained kNN neighbor and
\[
a_i^{\mathrm{mot}}
=
\left(
T^{-1}\sum_t
\|\widetilde{\boldsymbol\delta}_i(t)\|_2^2
\right)^{1/2}.
\]
Using the first- and second-order trajectory differences
\begin{align*}
\mathbf v_i^{\mathrm{mot}}(t)
&=
\Delta\boldsymbol\mu_i(t+1)-\Delta\boldsymbol\mu_i(t),\\
\mathbf b_i^{\mathrm{mot}}(t)
&=
\mathbf v_i^{\mathrm{mot}}(t+1)-\mathbf v_i^{\mathrm{mot}}(t),
\end{align*}
motion reliability is
\begin{align}
s_i^{\mathrm{mot}}
&=
\frac{
\operatorname{mean}_t\|\mathbf v_i^{\mathrm{mot}}(t)\|_2^2}
{
\operatorname{mean}_t\|\mathbf v_i^{\mathrm{mot}}(t)\|_2^2+
\operatorname{mean}_t\|\mathbf b_i^{\mathrm{mot}}(t)\|_2^2+\epsilon
},\\
u_i^{\mathrm{mot}}
&=
\frac{a_i^{\mathrm{mot}}}
{a_i^{\mathrm{mot}}+r_i}s_i^{\mathrm{mot}},
\qquad
u_{ij}^{\mathrm{mot}}
=
1-(1-u_i^{\mathrm{mot}})(1-u_j^{\mathrm{mot}}),\\
w_{ij}
&\leftarrow
\frac{w_{ij}+u_{ij}^{\mathrm{mot}}A_{ij}^{\mathrm{mot}}}
{1+u_{ij}^{\mathrm{mot}}}.
\label{eq:motion-reliability}
\end{align}
Motion updates only existing kNN edges. Definitions of the finite differences
and implementation details are provided in the supplementary material.

\subsection{Information-Theoretic Hierarchy}

For each harmonic branch, \(w_{ij}\) defines random-walk transition capacity.
Infomap minimizes hierarchical map-equation description length
\cite{rosvall2008maps}, producing labels
\(L_i^{(1)},\ldots,L_i^{(H)}\) from coarse to fine.

We define the opacity-weighted surface-area proxy
\begin{equation}
\omega_i=\alpha_i s_{i,(2)}s_{i,(3)},
\qquad
\max_R
\frac{\sum_{i\in R}\omega_i}
{\sum_i\omega_i}
\leq0.25,
\label{eq:mass-scale}
\end{equation}
where
\(s_{i,(1)}\leq s_{i,(2)}\leq s_{i,(3)}\)
are the ordered Gaussian semi-axes. For each harmonic branch, we select the
coarsest hierarchy level satisfying Eq.~\eqref{eq:mass-scale}. Between the
phase and invariant candidates, we retain the one with fewer regions,
breaking ties toward the invariant branch. The
candidate bank contains retained phase-hierarchy levels, this flat harmonic
candidate, and the four material partitions.

\subsection{Deterministic Point-Conditioned Retrieval}
\label{sec:retrieval}

Let the positive query be
\(p=(v_p,\mathbf x_p)\), where
\(\mathbf x_p\in\Omega_{v_p}\) and \(v_p\in\mathcal V\).
For each frozen partition \(q\), we retrieve the region whose rendered
categorical ID contains \(\mathbf x_p\), yielding persistent masks
\(P_q=\{P_q^v\}_{v\in\mathcal V}\). No target pixels are accessed after
query placement. Pairwise proposal agreement is
\begin{equation}
a(q,r)
=
\frac{
\sum_v |P_q^v\cap P_r^v|}
{\sum_v |P_q^v\cup P_r^v|}.
\label{eq:proposal-agreement}
\end{equation}

Consensus is balanced across partition families:
\begin{align}
A_f(q)
&=
\max_{r\neq q:\,\family(r)=f}a(q,r),\\
S(q)
&=
\frac{1}{|\mathcal F_q|}
\sum_{f\in\mathcal F_q}A_f(q)
-\beta c(q),
\qquad \beta=0.1,
\label{eq:consensus-score}
\end{align}
where \(c(q)\) is proposal coverage and \(\mathcal F_q\) contains families
with an available peer. Ties follow a fixed deterministic ordering.

\subsection{Common-Fate Refinement}

For dynamic scenes, consensus may select a material-consistent subregion.
Using \(T_r=12\) centered trajectories, define
\[
\mathbf d_i=
\operatorname{vec}\!\left(
\widetilde{\boldsymbol\delta}_i(t_1),\ldots,
\widetilde{\boldsymbol\delta}_i(t_{T_r})
\right),
\qquad
a_i^{\mathrm{cf}}=
\frac{\|\mathbf d_i\|_2}{\sqrt{T_r}},
\]
and
\(
\widehat{\mathbf d}_i=\mathbf d_i/\|\mathbf d_i\|_2
\),
with \(\widehat{\mathbf d}_i=\mathbf0\) for zero motion. For a region \(R\),
\begin{equation}
C(R)
=
\frac{
\left\|
\sum_{i\in R}
\omega_i a_i^{\mathrm{cf}}\widehat{\mathbf d}_i
\right\|_2}
{
\sum_{i\in R}\omega_i a_i^{\mathrm{cf}}
},
\label{eq:common-fate}
\end{equation}
when the denominator is nonzero, and \(C(R)=0\) otherwise.

We replace the consensus winner with the selected motion region when common
fate improves by more than \(0.05\). To control fragmentation, its
phase-hierarchy parent is selected when the coherence increase exceeds \(0.20\)
and parent coverage remains below \(0.40\). Dynamic evaluation always returns
one persistent region.

\subsection{Persistent-ID Rendering}

Region labels are stored on canonical Gaussian indices and follow their
deformation through time. Categorical rendering preserves these discrete IDs,
so the evaluated mask corresponds directly to the selected frozen Gaussian
region. Rendering implementation details are provided in the supplementary
material.
\section{Properties of the Selection Procedure}
\label{sec:analysis}

We establish four properties of the selection procedure: prediction is
independent of the evaluation mask conditioned on the point query, assigning
one maximum-agreement term to each proposal family prevents scale count from
dominating consensus, the common-fate score is a bounded measure of trajectory
agreement, and labels persist through rendering by construction. None of
these properties claims that intrinsic cues recover semantic object categories
in every scene. We detail mask independence and the common-fate bound below;
family balancing, temporal persistence, and complexity are detailed in the
supplementary material.

\paragraph{Mask independence.}
Let \(\Theta\) denote the frozen Gaussian and deformation checkpoints,
\(p=(v_p,\mathbf x_p)\) a positive point query (\S\ref{sec:retrieval}), and
\(\mathcal V\) a fixed set of rendered views. Every
partition is a function of \(\Theta\) and globally fixed constants; the
point query determines one integer label in each partition; and pairwise
agreement (Eq.~\eqref{eq:proposal-agreement}), coverage, and common fate
(Eq.~\eqref{eq:common-fate}) are all computed from rendered predictions and
checkpoint trajectories. None of these operations reads the evaluation mask
\(Y\), so for fixed \((\Theta,p,\mathcal V)\) the prediction is
invariant to \(Y\): the target is accessed only after selection, to compute
IoU. This does not mean the point-conditioned benchmark contains no supervision --
the simulated point query is derived from the target during evaluation -- only
that the target mask cannot affect proposal selection once the point query is
fixed.

\paragraph{Common fate is bounded.}
Equation~\eqref{eq:common-fate} is a weighted directional resultant. By the
triangle inequality it is bounded between 0 and 1, reaching 1 only when
every moving member of a region shares the same normalized trajectory and
shrinking as trajectories cancel. Common fate is therefore positive evidence
for a coherent moving extent, not a semantic score: a hand and a held object
can move together, while an articulated object can contain multiple
trajectories. The pre-fragmentation rule addresses the second case by
preferring the immediate parent when a large coherence gain coincides with
substantial mass removal.
\section{Experiments}
\label{sec:experiments}

\subsection{Experimental Setup}

\paragraph{Datasets.}
We evaluate on dynamic HyperNeRF \cite{park2021hypernerf} and Neu3D
\cite{li2022neu3d}. We report mean frame IoU, averaged within each object and
then macro-averaged over scenes. Because the datasets use different capture
regimes, we report them separately and use the overall macro only as a
summary.

\paragraph{Deterministic point query.}
Evaluation uses a single positive point inside the target object. To eliminate
operator variance, we select the annotated view with the largest visible target
area and place the point at its deepest interior pixel using a Euclidean
distance transform. This is the only stage at which the target affects
inference. After point placement, the target mask is hidden; the selector
retrieves one region from each frozen partition and applies
Eqs.~\eqref{eq:consensus-score} and \eqref{eq:common-fate}. The target mask is
then used only for scoring.

\paragraph{Prediction protocol.}
Each reported mIoU corresponds to the single prediction returned by the
deterministic selector. We do not use GT-selected proposals, hierarchy levels,
per-scene hyperparameters, target-selected region unions, or Boolean
combinations selected by target overlap. Target overlap is used only for
evaluation.

\paragraph{Implementation details.}
All partitions are generated once from the frozen reconstruction. We use
\(k=20\), opacity threshold \(0.05\), and material graph resolutions
\(\{0.004,0.008,0.015,0.060\}\). The dual harmonic graphs use scene-median cue
calibration, one deterministic Infomap trial, and \(T=20\) deformation samples.
Common-fate refinement uses \(T_r=12\), coverage penalty \(\beta=0.1\), motion
override margin \(0.05\), fragmentation jump \(0.20\), and maximum parent
coverage \(0.40\). We use \(\epsilon=10^{-8}\) for all numerical stabilizers.
All constants are fixed across scenes. No segmentation optimizer, foundation
model, or feature network is used.

\subsection{Dynamic Segmentation Results}

\begin{table*}[t]
\centering
\small
\renewcommand{\arraystretch}{1.08}

\resizebox{0.85\textwidth}{!}{%
\begin{tabular*}{\textwidth}{@{\extracolsep{\fill}}l c c c c c c c}
\toprule
\multicolumn{8}{c}{\textbf{Neu3D \cite{li2022neu3d}}} \\
\midrule
Method & MF & Coffee & Spinach & Beef & Flame & Sear & Avg. \\
\midrule

SA4D \cite{ji2024sa4d}
& \xmark
& \secondbest{.858}
& .899
& .865
& \secondbest{.890}
& .905
& .883 \\

Gaussian Grouping \cite{ye2024gg}
& \xmark
& .830
& .897
& \best{.951}
& .828
& \best{.926}
& .886 \\

TRASE \cite{li2026trase}
& \xmark
& \best{.912}
& \secondbest{.913}
& \secondbest{.910}
& .872
& .904
& \best{.902} \\

\midrule

\method (ours)
& \cmark
& .834
& \best{.932}
& .907
& \best{.911}
& \secondbest{.913}
& \secondbest{.899} \\

\bottomrule
\end{tabular*}
}

\vspace{0.65em}

\resizebox{0.85\textwidth}{!}{%
\begin{tabular*}{\textwidth}{@{\extracolsep{\fill}}l c *{11}{c}}
\toprule
\multicolumn{13}{c}{\textbf{HyperNeRF \cite{park2021hypernerf}}} \\
\midrule
Method & MF & Am. & Chk. & Lem. & Espr. & Hand & Key. & Mitts & Ban. & Ck. & Ch. & Avg. \\
\midrule

SA4D \cite{ji2024sa4d}
& \xmark
& .849
& .827
& \secondbest{.816}
& .451
& .706
& .845
& .726
& .721
& .835
& .647
& .742 \\

Gaussian Grouping \cite{ye2024gg}
& \xmark
& \secondbest{.898}
& \best{.944}
& .715
& .497
& .887
& \best{.916}
& \secondbest{.742}
& \best{.910}
& \best{.913}
& \best{.920}
& \secondbest{.834} \\

TRASE \cite{li2026trase}
& \xmark
& .814
& \secondbest{.931}
& \best{.880}
& \secondbest{.716}
& \best{.926}
& \secondbest{.870}
& \best{.930}
& \secondbest{.878}
& \secondbest{.858}
& \secondbest{.860}
& \best{.866} \\

\midrule

\method (ours)
& \cmark
& \best{.912}
& .599
& .185
& \best{.785}
& \secondbest{.905}
& .437
& .576
& .174
& .421
& .667
& .566 \\

\bottomrule
\end{tabular*}
}

\caption{\textbf{Dynamic segmentation results.}
Point-conditioned mIoU on Neu3D and HyperNeRF.
MF denotes mask-free segmentation: \xmark indicates methods that use 2D mask
supervision, while \cmark indicates \method, which uses no such masks.
\method reaches within 0.003 mIoU of TRASE on Neu3D without mask supervision
and returns one persistent region per point query. After point placement, the
GT mask does not affect region selection. Best results are \textbf{bold}; second
best are \underline{underlined}.}
\label{tab:main-results}
\end{table*}

\paragraph{Multi-view Neu3D.}
\method reaches 0.899 mIoU on Neu3D, within 0.003 of TRASE's 0.902 mIoU,
without mask supervision, external visual features, or segmentation training.
The multi-view capture regime provides stronger constraints on the reconstructed
Gaussian field, allowing geometry, SH appearance, and common fate to approach
mask-supervised segmentation performance.

\paragraph{Monocular HyperNeRF.}
\method reaches 0.566 mIoU on HyperNeRF. The largest failures occur in
cut-lemon and slice-banana, where the benchmark target combines components that
the intrinsic hierarchy separates. Americano, espresso, hand, and torchocolate
show substantially stronger recovery using only intrinsic Gaussian cues.

\paragraph{Overall.}
Across both datasets, \method obtains a macro-average of 0.677 mIoU. The gap
between Neu3D and HyperNeRF indicates that intrinsic segmentation quality
depends strongly on the constraints provided by the reconstruction regime.

\subsection{Static-Scene Transfer}
\label{sec:static-generalization}

We evaluate qualitative transfer on two static scenes, NVOS Orchids
\cite{goel2023nvos} and the LERF-Extended Bouquet scene \cite{kerr2023lerf},
containing repeated structures, thin geometry, occlusion, and background
clutter.

\paragraph{Visualization protocol.}
For full-scene visualization only, we apply a fixed, GT-independent selection
rule: the coarsest partition before its largest region exceeds 65\% of
assigned Gaussians. This rule is separate from the one-region selector used
for quantitative evaluation. The exact procedure, including how nested
regions are retained across partition families, is provided in the
supplementary material.

Figure~\ref{fig:static-generalization} shows that the intrinsic partition bank
separates individual blossoms, leaves, and stems while maintaining coherent
regions across cluttered backgrounds. No masks, semantic encoders, or
scene-specific tuning are used. Quantitative evaluation on broader static
datasets and reconstruction families remains future work.

\begin{figure}[t]
    \centering
    \includegraphics[width=\columnwidth]{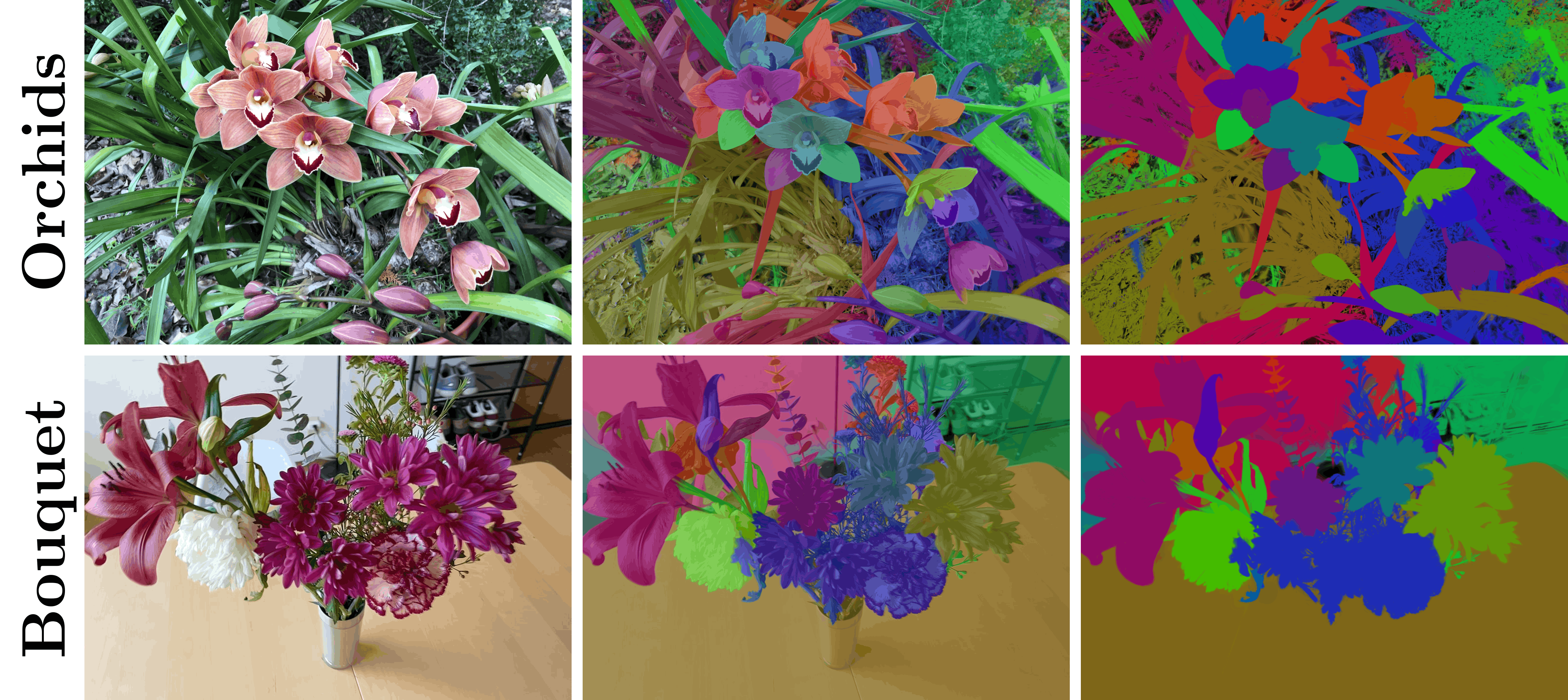}
    \caption{\textbf{Qualitative transfer to static scenes.}
    Orchids and Bouquet contain repeated flowers, thin leaves and stems,
    occlusion, and background clutter. From left to right: RGB, the frozen
    full-scene partition at 65\% opacity over RGB, and the same categorical
    partition without RGB. The visualization rule operates only on the
    intrinsic partition bank and is independent of ground truth.}
    \label{fig:static-generalization}
\end{figure}

\subsection{Ablation Studies}

\begin{table}[t]
\centering
\small
\setlength{\tabcolsep}{3.0pt}
\resizebox{0.85\columnwidth}{!}{%
\begin{tabular}{lccc}
\toprule
Variant & HyperNeRF & Neu3D & Overall \\
\midrule
No common fate & .405 & .899 & .570 \\
No pre-fragmentation stop & .520 & .899 & .647 \\
No coverage penalty & .529 & .899 & .653 \\
Candidate-mean consensus & .562 & .793 & .639 \\
Full method & \best{.566} & \best{.899} & \best{.677} \\
\bottomrule
\end{tabular}
}
\caption{\textbf{Selector ablations.}
All variants use identical frozen partitions and point queries.
Best results are \textbf{bold}; second best are \underline{underlined}.}
\label{tab:selector-ablation}
\end{table}

Common fate provides the largest gain on HyperNeRF, increasing mIoU from
0.405 to 0.566 while leaving Neu3D unchanged. Fragmentation control and the
coverage penalty improve the overall macro from 0.647 and 0.653 to 0.677,
respectively. Replacing family-balanced consensus with candidate averaging
reduces Neu3D from 0.899 to 0.793, showing the importance of balancing
partition families rather than individual hierarchy levels. All variants
return one region, and target overlap is used only for scoring. Additional
cue-family ablations and point-location sensitivity are provided in the
supplementary material.

\subsection{Efficiency}

\paragraph{Runtime setup.}
We rerun each scene from its trained Gaussian checkpoint on a single consumer
RTX 5080 with 16 GB VRAM using PyTorch 2.8 and CUDA 12.8. Preprocessing
constructs the four material partitions and the phase and invariant harmonic
branches sequentially. Wall time excludes Gaussian reconstruction training,
categorical visualization rendering, and metric computation. Preprocessing
averages 190.4 seconds per scene with 3.01 GiB mean peak allocated VRAM.
Each timed run records the checkpoint hash, stage-level timings, hardware
metadata, and reproduced point-conditioned mIoU.

\paragraph{Comparison with TRASE.}
Using the same per-scene HyperNeRF stage accounting,
TRASE~\cite{li2026trase} requires 2136.8 seconds per scene for SAM automatic
mask generation and 122.9 seconds for feature rendering, totaling 2259.6
seconds. GS-IntSeg requires 150.3 seconds per scene to construct its four
material partitions and two harmonic branches from a trained checkpoint,
giving a \(15.0{\times}\) reduction in checkpoint-to-segmentation preprocessing
time. The comparison excludes TRASE's approximately 20k iterations of
contrastive feature training and excludes categorical visualization rendering
for GS-IntSeg. No corresponding TRASE timing is available for Neu3D; we
therefore report no Neu3D speedup.
\section{Discussion}
\label{sec:discussion}

The experiments show that the amount of recoverable object structure depends
strongly on the reconstruction regime. On multi-view Neu3D, intrinsic Gaussian
geometry, radiance, and motion are sufficient to reach 0.899 mIoU, close to
mask-supervised methods. On monocular HyperNeRF, the same cues are less
sufficient because reconstruction ambiguity and object interaction weaken the
correspondence between physical coherence and semantic object extent.

This distinction clarifies what intrinsic Gaussian segmentation can and cannot
recover. Geometry, appearance continuity, and common motion provide strong
evidence for physically coherent regions, but they do not uniquely determine
semantic grouping. A hand and a held object may move together, while one
articulated object may contain several distinct motion patterns. The remaining
HyperNeRF gap therefore reflects ambiguity in semantic extent rather than the
absence of useful structure in the Gaussian representation.

\paragraph{Scope of mask-free segmentation.}
\method does not use segmentation masks to construct partitions, compute
affinities, rank candidates, or refine the selected region. Ground-truth masks
are used only to place the deterministic positive point and compute evaluation
metrics. The Gaussian reconstruction itself is trained from RGB observations,
so ``mask-free'' refers specifically to the segmentation stage.

\paragraph{Implications.}
The Gaussian-only setting isolates object information already encoded by the
reconstruction before external semantic supervision is introduced. This
provides a reference point for determining when external features are actually
necessary. A natural extension is to combine intrinsic geometry, SH appearance,
and common fate with self-supervised visual descriptors while preserving direct
segmentation over persistent Gaussians and avoiding 2D mask supervision.

\section{Limitations}

Intrinsic grouping is class agnostic and therefore does not always coincide
with semantic object boundaries. Common motion can merge interacting objects,
while articulation can split a single object into multiple regions. SH
coefficients encode radiance rather than semantics, so illumination,
specularity, exposure, and reconstruction errors can introduce appearance
boundaries unrelated to object identity. The center-space kNN graph can also
connect nearby contacting surfaces or underrepresent thin structures.

Quantitative evaluation is currently limited to HyperNeRF and Neu3D dynamic
reconstructions. The static-scene experiments are qualitative, and broader
evaluation across static benchmarks, reconstruction methods, and capture
conditions is required to establish generalization.

\section{Conclusion}

We introduced the first intrinsic, mask-free formulation for segmenting frozen
Gaussian scenes using geometry, opacity, spherical-harmonic appearance, and
deformation trajectories. On multi-view Neu3D, these cues reach 0.899 mIoU
without segmentation masks, external visual features, or segmentation
training. Performance decreases to 0.566 mIoU on monocular HyperNeRF, where
physical grouping alone is less sufficient to determine semantic object
extent.

These results show that a substantial segmentation signal is already encoded
in trained Gaussian representations. Intrinsic Gaussian segmentation therefore
provides a direct baseline for studying how much object structure can be
recovered before introducing external semantic supervision, and a foundation
for efficient 3D and 4D segmentation with progressively richer cues.

\clearpage
{
    \small
    \bibliographystyle{ieeenat_fullname}
    \bibliography{main}

@inproceedings{li2026trase,
  title={{TRASE}: Tracking-free {4D} Segmentation and Editing},
  author={Li, Yun-Jin and Gladkova, Mariia and Xia, Yan and Cremers, Daniel},
  booktitle={Proc. International Conference on 3D Vision (3DV)},
  year={2026}
}

@inproceedings{ye2024gg,
  title={Gaussian Grouping: Segment and Edit Anything in {3D} Scenes},
  author={Ye, Mingqiao and Danelljan, Martin and Yu, Fisher and Ke, Lei},
  booktitle={Proc. European Conference on Computer Vision (ECCV)},
  year={2024}
}

@article{ji2024sa4d,
  title={{SA4D}: Segment Any {4D} {Gaussians}},
  author={Ji, Shengxiang and Wu, Guanjun and Fang, Jiemin and Cen, Jiazhong
          and Yi, Taoran and Liu, Wenyu and Tian, Qi and Wang, Xinggang},
  journal={arXiv preprint arXiv:2407.04504},
  year={2024}
}

@inproceedings{cen2024saga,
  title={{SAGA}: Segment Any {3D} {Gaussians}},
  author={Cen, Jiazhong and Fang, Jiemin and Yang, Chen and Xie, Lingxi
          and Zhang, Xiaopeng and Shen, Wei and Tian, Qi},
  booktitle={Proc. AAAI Conference on Artificial Intelligence (AAAI)},
  year={2025}
}

@inproceedings{bae2024dgd,
  title={{DGD}: Dynamic {3D} {Gaussians} Distillation},
  author={Labe, Isaac and Issachar, Noam and Lang, Itai and Benaim, Sagie},
  booktitle={Proc. European Conference on Computer Vision (ECCV)},
  year={2024}
}

@inproceedings{wu2024opengaussian,
  title={{OpenGaussian}: Towards Point-Level {3D} {Gaussian}-based
         Open Vocabulary Understanding},
  author={Wu, Yanmin and Meng, Jiarui and Li, Haijie and Wu, Chenming
          and Shi, Yahao and Cheng, Xinhua and Zhao, Chen and Feng, Haocheng
          and Ding, Errui and Wang, Jingdong and Zhang, Jian},
  booktitle={Advances in Neural Information Processing Systems (NeurIPS)},
  year={2024}
}

@article{barhdadi2026_4dsync,
  title={{4D} Synchronized Fields: Motion-Language {Gaussian} Splatting
         for Temporal Scene Understanding},
  author={Barhdadi, Mohamed Rayan and Abdaljalil, Samir and Khanbayov, Rasul
          and Serpedin, Erchin and Kurban, Hasan},
  journal={arXiv preprint arXiv:2603.14301},
  year={2026}
}

@inproceedings{qin2024langsplat,
  title={{LangSplat}: {3D} Language {Gaussian} Splatting},
  author={Qin, Minghan and Li, Wanhua and Zhou, Jiawei and Wang, Haoqian
          and Pfister, Hanspeter},
  booktitle={Proc. IEEE/CVF Conference on Computer Vision and Pattern
             Recognition (CVPR)},
  year={2024}
}

@inproceedings{zhou2024feature3dgs,
  title={Feature {3DGS}: Supercharging {3D} {Gaussian} Splatting to Enable
         Distilled Feature Fields},
  author={Zhou, Shijie and Chang, Haoran and Jiang, Sicheng and Fan, Zhiwen
          and Zhu, Zehao and Xu, Dejia and Chari, Pradyumna and You, Suya
          and Wang, Zhangyang and Kadambi, Achuta},
  booktitle={Proc. IEEE/CVF Conference on Computer Vision and Pattern
             Recognition (CVPR)},
  year={2024}
}

@inproceedings{kerr2023lerf,
  title={{LERF}: Language Embedded Radiance Fields},
  author={Kerr, Justin and Kim, Chung Min and Goldberg, Ken and
          Kanazawa, Angjoo and Tancik, Matthew},
  booktitle={Proc. IEEE/CVF International Conference on Computer Vision (ICCV)},
  year={2023}
}

@inproceedings{jain2024gaussiancut,
  title={{GaussianCut}: Interactive Segmentation via Graph Cut for
         {3D} {Gaussian} Splatting},
  author={Jain, Umangi and Mirzaei, Ashkan and Gilitschenski, Igor},
  booktitle={Advances in Neural Information Processing Systems (NeurIPS)},
  year={2024}
}

@inproceedings{shen2024flashsplat,
  title={{FlashSplat}: {2D} to {3D} {Gaussian} Splatting Segmentation Solved
         Optimally},
  author={Shen, Qiuhong and Yang, Xingyi and Wang, Xinchao},
  booktitle={Proc. European Conference on Computer Vision (ECCV)},
  year={2024}
}

@article{kirillov2023segany,
  title={Segment Anything},
  author={Kirillov, Alexander and Mintun, Eric and Ravi, Nikhila and Mao, Hanzi and Rolland, Chloe and Gustafson, Laura and Xiao, Tete and Whitehead, Spencer and Berg, Alexander C. and Lo, Wan-Yen and Doll{\'a}r, Piotr and Girshick, Ross},
  journal={arXiv:2304.02643},
  year={2023}
}

@inproceedings{ravi2024sam2,
  title={{SAM} 2: Segment Anything in Images and Videos},
  author={Ravi, Nikhila and Gabeur, Valentin and Hu, Yuan-Ting and Hu, Ronghang
          and Ryali, Chaitanya and Ma, Tengyu and Khedr, Haitham and R{\"a}dle, Roman
          and Rolland, Chloe and Gustafson, Laura and Mintun, Eric and Pan, Junting
          and Alwala, Kalyan Vasudev and Carion, Nicolas and Wu, Chao-Yuan
          and Girshick, Ross and Doll{\'a}r, Piotr and Feichtenhofer, Christoph},
  booktitle={Proc. International Conference on Learning Representations (ICLR)},
  year={2025}
}

@inproceedings{carion2025sam3,
  title={{SAM} 3: Segment Anything with Concepts},
  author={Carion, Nicolas and Gustafson, Laura and Hu, Yuan-Ting and
          Debnath, Shoubhik and Hu, Ronghang and Suris Coll-Vinent, Didac and
          Ryali, Chaitanya and Alwala, Kalyan Vasudev and Khedr, Haitham and
          Huang, Andrew and Lei, Jie and Ma, Tengyu and Guo, Baishan and
          Kalla, Arpit and Marks, Markus and Greer, Joseph and Wang, Meng and
          Sun, Peize and R{\"a}dle, Roman and Afouras, Triantafyllos and others},
  booktitle={Proc. International Conference on Learning Representations (ICLR)},
  year={2026}
}

@article{park2021hypernerf,
  title={{HyperNeRF}: A Higher-Dimensional Representation for
         Topologically Varying Neural Radiance Fields},
  author={Park, Keunhong and Sinha, Utkarsh and Hedman, Peter and
          Barron, Jonathan T. and Bouaziz, Sofien and Goldman, Dan B. and
          Martin-Brualla, Ricardo and Seitz, Steven M.},
  journal={ACM Transactions on Graphics},
  volume={40}, number={6}, pages={1--12},
  year={2021}
}

@inproceedings{li2022neu3d,
  title={Neural {3D} Video Synthesis from Multi-view Video},
  author={Li, Tianye and Slavcheva, Mira and Zollhoefer, Michael and
          Green, Simon and Lassner, Christoph and Kim, Changil and
          Schmidt, Tanner and Lovegrove, Steven and Goesele, Michael and
          Newcombe, Richard and Lv, Zhaoyang},
  booktitle={Proc. IEEE/CVF Conference on Computer Vision and Pattern
             Recognition (CVPR)},
  year={2022}
}

@article{kerbl2023gs,
  title={{3D} {Gaussian} Splatting for Real-Time Radiance Field Rendering},
  author={Kerbl, Bernhard and Kopanas, Georgios and Leimk{\"u}hler, Thomas
          and Drettakis, George},
  journal={ACM Transactions on Graphics},
  volume={42}, number={4},
  year={2023}
}

@inproceedings{wu2024gs4d,
  title={{4D} {Gaussian} Splatting for Real-Time Dynamic Scene Rendering},
  author={Wu, Guanjun and Yi, Taoran and Fang, Jiemin and Xie, Lingxi and
          Zhang, Xiaopeng and Wei, Wei and Liu, Wenyu and Tian, Qi and
          Wang, Xinggang},
  booktitle={Proc. IEEE/CVF Conference on Computer Vision and Pattern
             Recognition (CVPR)},
  year={2024}
}

@inproceedings{yang2024deformable3dgs,
  title={Deformable {3D} {Gaussians} for High-Fidelity Monocular Dynamic
         Scene Reconstruction},
  author={Yang, Ziyi and Gao, Xinyu and Zhou, Wen and Jiao, Shaohui and
          Zhang, Yuqing and Jin, Xiaogang},
  booktitle={Proc. IEEE/CVF Conference on Computer Vision and Pattern
             Recognition (CVPR)},
  year={2024}
}

@article{shi2000ncut,
  title={Normalized Cuts and Image Segmentation},
  author={Shi, Jianbo and Malik, Jitendra},
  journal={IEEE Transactions on Pattern Analysis and Machine Intelligence},
  volume={22}, number={8}, pages={888--905},
  year={2000}
}

@article{traag2019leiden,
  title={From {Louvain} to {Leiden}: Guaranteeing Well-Connected Communities},
  author={Traag, Vincent A. and Waltman, Ludo and van Eck, Nees Jan},
  journal={Scientific Reports},
  volume={9}, number={1}, pages={5233},
  year={2019}
}

@article{spelke1990principles,
  title={Principles of Object Perception},
  author={Spelke, Elizabeth S.},
  journal={Cognitive Science},
  volume={14}, number={1}, pages={29--56},
  year={1990}
}

@article{spelke2000core,
  title={Core Knowledge},
  author={Spelke, Elizabeth S.},
  journal={American Psychologist},
  volume={55},
  number={11},
  pages={1233--1243},
  year={2000}
}

@book{koffka1935gestalt,
  title={Principles of Gestalt Psychology},
  author={Koffka, Kurt},
  publisher={Harcourt, Brace and Company},
  year={1935}
}

@inproceedings{li20254dlangsplat,
  title     = {4D LangSplat: 4D Language Gaussian Splatting via Multimodal Large Language Models},
  author    = {Li, Wanhua and Zhou, Renping and Zhou, Jiawei and Song, Yingwei and Herter, Johannes and Qin, Minghan and Huang, Gao and Pfister, Hanspeter},
  booktitle = {Proceedings of the IEEE/CVF Conference on Computer Vision and Pattern Recognition (CVPR)},
  pages     = {22001--22011},
  year      = {2025}
}

@article{li2025langsplatv2,
  title   = {LangSplatV2: High-dimensional 3D Language Gaussian Splatting with 450+ FPS},
  author  = {Li, Wanhua and Zhao, Yujie and Qin, Minghan and Liu, Yang and Cai, Yuanhao and Gan, Chuang and Pfister, Hanspeter},
  journal = {Advances in Neural Information Processing Systems (NeurIPS)},
  year    = {2025}
}

@article{rosvall2008maps,
  title   = {Maps of Random Walks on Complex Networks Reveal Community Structure},
  author  = {Rosvall, Martin and Bergstrom, Carl T.},
  journal = {Proceedings of the National Academy of Sciences},
  volume  = {105},
  number  = {4},
  pages   = {1118--1123},
  year    = {2008}
}

@inproceedings{marrie2025ludvig,
  title     = {{LUDVIG}: Learning-Free Uplifting of {2D} Visual Features to
               Gaussian Splatting Scenes},
  author    = {Marrie, Juliette and Menegaux, Romain and Arbel, Michael and
               Larlus, Diane and Mairal, Julien},
  booktitle = {Proceedings of the IEEE/CVF International Conference on
               Computer Vision (ICCV)},
  pages     = {7440--7450},
  year      = {2025}
}

@article{split4d2025,
  title={{Split4D}: Decomposed {4D} Scene Reconstruction Without Video Segmentation},
  author={Hu, Yongzhen and Yang, Yihui and Lin, Haotong and Wang, Yifan and
          Dong, Junting and Deng, Yifu and Zhu, Xinyu and Jia, Fan and
          Bao, Hujun and Zhou, Xiaowei and Peng, Sida},
  journal={arXiv preprint arXiv:2512.22745},
  year={2025}
}

@inproceedings{goel2023nvos,
  title     = {Interactive Segmentation of Radiance Fields},
  author    = {Goel, Rahul and Sirikonda, Dhawal and Saini, Saurabh and
               Narayanan, P.J.},
  booktitle = {Proceedings of the IEEE/CVF Conference on Computer Vision
               and Pattern Recognition (CVPR)},
  pages     = {4201--4211},
  year      = {2023}
}
}


\end{document}